# A Spatial-Temporal Dual-Mode Mixed Flow Network for Panoramic Video Salient Object Detection

Xiaolei Chen, Pengcheng Zhang, Zelong Du, Ishfaq Ahmad, *Fellow, IEEE*

**Abstract**—Salient object detection (SOD) in panoramic video is still in the initial exploration stage. The indirect application of 2D video SOD method to the detection of salient objects in panoramic video has many unmet challenges, such as low detection accuracy, high model complexity, and poor generalization performance. To overcome these hurdles, we design an Inter-Layer Attention (ILA) module, an Inter-Layer weight (ILW) module, and a Bi-Modal Attention (BMA) module. Based on these modules, we propose a Spatial-Temporal Dual-Mode Mixed Flow Network (STDMMF-Net) that exploits the spatial flow of panoramic video and the corresponding optical flow for SOD. First, the ILA module calculates the attention between adjacent level features of consecutive frames of panoramic video to improve the accuracy of extracting salient object features from the spatial flow. Then, the ILW module quantifies the salient object information contained in the features of each level to improve the fusion efficiency of the features of each level in the mixed flow. Finally, the BMA module improves the detection accuracy of STDMMF-Net. A large number of subjective and objective experimental results testify that the proposed method demonstrates better detection accuracy than the state-of-the-art (SOTA) methods. Moreover, the comprehensive performance of the proposed method is better in terms of memory required for model inference, testing time, complexity, and generalization performance.

## 1. INTRODUCTION

The main goal of video salient object detection (SOD) is to find the most eye-catching objects in videos [1],[2],[3]. The temporal information provided by video makes the task of video SOD more difficult, because compared with spatial salient object information, the human visual system is more likely to be attracted by temporal salient information[4],[5],[6]. For example, for an image, the human eye will pay attention to the salient objects that are different from the surrounding environment based on texture or color. However, for a video, the human eye will also be attracted by moving objects [7].

With the rapid proliferation of VR and AR technology, SOD in panoramic video has become one of the new key problems in the field of computer vision in the past three years, because it can improve the editing, compression, and transmission efficiency of panoramic video. SOD in the panoramic video is more difficult than that in the traditional 2D video. The acquisition of panoramic video also requires professional equipment and the labeling requires a lot of effort. Currently, the publicly available data sets that can be used for SOD in panoramic video are very limited; only two data sets, SHD360[8] and ASOD60K[9] are available.

Second, because the relevant methods and technologies for SOD in panoramic video are still in the exploration stage, and the work for reference is very limited.

There is scant literature on this topic. The work reported in [10] proposes an audio-assisted SOD model for panoramic video, which utilizes the spatial correspondence between audio and video to accomplish the SOD task for panoramic video . However, the introduction of audio for SOD in panoramic video limits its application scenarios of [10]. For example, by default there is no audio information available in a soundless environment and introducing audio in a noisy environment instead can cause a degradation of detection performance. To facilitate the distinction, our work refers to [10] as the direct panoramic video SOD method based on audio and video information.

The SOD methods for traditional 2D video can be indirectly applied to SOD in panoramic video. This paper refers to these kinds of methods as the indirect panoramic video SOD methods. The state-of-the-art indirect methods in recent years are as follows:

Wang *et al.* [11] proposed a fully convolutional neural network, which inputs consecutive frames and corresponding single frames of video as spatial saliency information and temporal salient information respectively, so as to quickly complete video SOD. Song *et al.* [12] proposed a recurrent network architecture based on pyramid-expanded convolution to extract features from two adjacent frames for rapid video SOD. Shokri *et al.* [13] used depth non-local neural networks to detect video salient objects frame by frame and then determine salient objects by calculating global dependencies. These three methods detect salient objects through single or consecutive video frames. Because panoramic video has high resolution, it is

This work is supported by the National Natural Science Foundation of China (61967012) (Corresponding author: Xiaolei Chen)

Xiaolei Chen is currently an associate professor in the College of Electrical and Information Engineering at Lanzhou University of Technology, China (e-mail: chenxl703@lut.edu.cn).

Pengcheng Zhang is currently a graduate student in the College of Electrical and Information Engineering at Lanzhou University of Technology, China (e-mail: 2072986368@qq.com).

Zelong Du is currently a graduate student in the College of Electrical and Information Engineering at Lanzhou University of Technology, China (e-mail: 1132911812@qq.com).

Ishfaq Ahmad Ishfaq Ahmad is currently a professor of Computer Science and Engineering Department at the University of Texas at Arlington, USA(e-mail: iahmad@cse.uta.edu

necessary to downsample them and then extract features due to the limitations of current computing power and other factors. However, downsampling itself may lead to the loss of temporal information of panoramic video, which will seriously reduce the detection efficiency of SOD in panoramic video.

Zhao *et al.* [14] proposed a motion-sensing memory network for rapid detection of video salient objects, which requires a large number of complex preparations before detection. Ehteshami *et al.* [15] proposed a non-uniform sampling input technique, which detects salient objects in the first frame of video at the original resolution, and the obtained salient map is used to guide the resampling of the second frame. However, this method cannot accurately detect fast moving objects or small objects. In addition, because panoramic video has a wide angle of view, the background usually occupies a large proportion, and the salient objects occupy a small proportion of the whole frame. Therefore, using this method to perform SOD in panoramic video cannot fully leverage its advantages.

Jiao *et al.* [16] proposed guidance and teaching network, which uses implicit guidance and explicit teaching to independently extract effective spatial and temporal clues at feature level and decision level respectively, and then complete SOD. Su *et al.* [17] proposed a converter framework based on group segmentation, which is used for Co-Segmentation (CoS), Co-Saliency Detection (CoSD) and video saliency target detection to find Co-Occurrent Objects. In this framework, image features are regarded as patch markers, and the long dependence between features is calculated by self-attention mechanism. When these two methods are used for SOD in panoramic video, the accuracy needs to be improved.

Zhang *et al.* [18] proposed a dynamic context-aware filtering network, which extracts the multi-layer features of three consecutive video frames, calculates the relative relationship between adjacent frames, and transmits them frame by frame to obtain salient objects. Lu *et al.* [19] proposed a deep collaborative three-mode network, which uses a multi-mode attention module to simulate the dependence between main mode and two auxiliary modes, suppresses noise through a fusion module, and obtains the detection results of video salient objects. Tang *et al.* [20] proposed an adaptive local global refinement framework, which refines feature fusion at different scales. Chen *et al.* [21] proposed a video SOD method based on motion quality perception, frames with high-quality motion information are selected from original video frames for model training. The above models have high computational load, therefore the detection time will be a great challenge for the SOD in panoramic video.

Zhang *et al.* [22] proposed a video SOD method based on depth map quality, which takes depth features as weights and controls the fusion of spatial features and depth features. Liu *et al.* [23] proposed a dynamic spatial-temporal network according to the complementary effect of spatial-temporal information, which facilitates dynamic complementary aggregation of spatial-temporal features through a cross attentive aggregation procedure. Chen *et al.* [24] proposed a spatial-temporal network, which uses spatial branches to periodically refine temporal branches in a multi-scale manner, so as to achieve performance improvement. Li *et al.* [25] proposed a motion-guided attention network for video SOD, the network is composed of two sub-networks, one sub-network for SOD in still images, and the other for motion saliency detection in optical flow images, and their proposed attention modules bridge these two sub-networks. The above methods focus on specific data sets, and the detection effect is poor when using panoramic video data sets for detection. At the same time, the generalization performance of these models has room for improvement.

In summary, indirect panoramic video SOD methods have some problems, such as low detection accuracy, high memory required for model reasoning, long test time, high model complexity, and poor generalization performance. To solve these problems, in this paper a spatial-temporal dual-mode mixed flow network (STDMMF-Net) is proposed to detect salient objects in panoramic video. As far as we know, this is the second SOD method for panoramic video after literature [10], and different from literature [10], this method is a direct SOD method for panoramic video based on video information, and applies to a wider range of scenarios than literature [10].

The main contributions and innovations of this paper are as follows:

(1) The STDMMF-Net is proposed, which uses spatial flow and optical flow as inputs, extracts salient object features at the same time, and integrates the two flows to detect salient objects in panoramic video. Its comprehensive performance is better than the existing seven SOTA methods.

(2) An interlayer attention (ILA) module is designed to calculate the attention between adjacent level features of panoramic video consecutive frames in the spatial flow, so as to improve the accuracy of extracting salient object features from the spatial flow.

(3) An interlayer weight (ILW) module is presented, which calculates the weight of each level feature according to the amount of salient object information contained in each of the five levels of spatial and temporal flows, and quantifies the salient object information contained in each level feature.

(4) A bimodal attention (BMA) module is proposed to calculate the dual-mode attention oriented to the output features of each level of spatial flow and time flow, and improve the detection accuracy of the STDMMF-Net.

## 2. The Proposed Model

### 2.1. Spatial-Temporal Dual-mode Mixed Flow Network (STDMMF-Net)

The overall structure of STDMMF-Net for panoramic video SOD is shown in Fig. 1. The overall structure is composed of spatial flow, time flow, and mixed flow, which are represented by blue, and green arrows respectively in Fig. 1. The overall working principle is as follows:

The spatial flow takes consecutive panoramic video frames as inputs. Firstly, the ResNet34 is used to extract the five input hierarchical features f1-f5, and the attention A1-A4 among adjacent layers of f1-f5 is calculated by the ILA module. At the same time, the features of panoramic video consecutive frames are extracted by layer 1 of ResNet34, and then the feature map L1-S is obtained under the action of interlayer attention A1. L1-S extracts feature x1 through layer 2 of ResNet34, and x1 obtains feature map L2-S under the action of interlayer attention A2; Feature maps L3-S and L4-S are obtained by the same method. Finally, L4-S was pooled by the Atrous Spatial Pyramid Pooling (ASPP) to obtain the feature map L5-S.

The time flow takes the optical flow corresponding to panoramic video as input, and uses the ResNet34 to extract five hierarchical features L1-T, L2-T, L3-T, L4-T and L5-T of optical flow frames.

The output features Li-S and Li-T (i=1,2,3,4,5) of space flow and time flow are simultaneously used as the input of mixed flow. Firstly, the interlayer weight is calculated by the ILW module. After convolution, batch normalization and rectified linear unit (CBR) operations, the spatial flow high level features L4-S and L5-S and the temporal flow high level features L4-T and L5-T, together with interlayer weight, are used as inputs of the BMA module to obtain bimodal attention (Bi-att). The corresponding hierarchical features of space flow and time flow are used under the action of interlayer weight to obtain the five hierarchical features of mixed flow Mix1-Mix5. These five hierarchical features are fused with Bi-att through multi-level feature fusion to obtain the final salient object detection results.

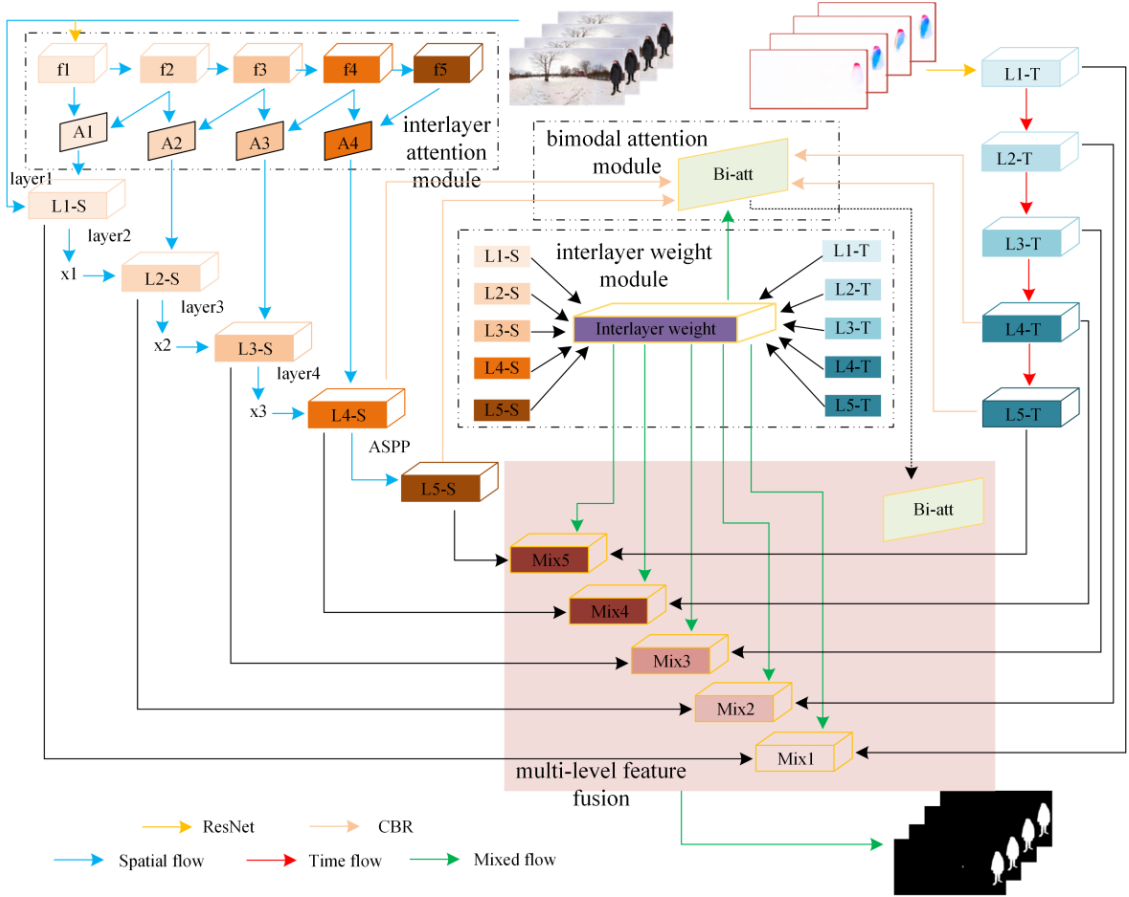

**Fig. 1.** Overall structure diagram of STDMMF-Net

## 2.2. Spatial flow processing

Panoramic video consecutive frames contain abundant salient object information, which is very important for high quality SOD in panoramic video. The interlayer attention module and the feature generation process of each layer are described below.

### 2.2.1. Interlayer attention module

The multi-level features output by the ResNet34 have different spatial and channel scales, and contain different levels of semantic and detailed information. This paper calculates the interlayer attention from two adjacent layers. The structure of the ILA module is shown in Fig. 2. The low level features fi (i=1,2,3,4) with the scale of (Ci, hi, wi) are convolved by the combination of different convolution kernels, and the salient object information contained in the low level feature is extracted with different receptive fields, and the number of channels is compressed to 64. The size of convolution kernels includes 1×1, 1×3, 3×1, 1×5, 5×1, 1×7, 7×1, and 3×3. The four features Xi-0, Xi-1, Xi-2, and Xi-3 are obtained by parallel convolution of fi (i=1,2,3,4) in the four combinations shown in Fig. 2. These four features are concatenated in the channel direction to obtain the concatenated feature Xi-cat with scale (4×64,hi,wi), and then the channels number of Xi-cat is compressed to 64 by convolution and batch normalization (BN). The result is added with Xi-0 to prevent module performance degradation. Finally, the Relu activation operation is performed to obtain a feature Fi with the scale of (64,hi,wi).

The high level feature fi+1 with scale (Ci+1,hi+1,wi+1) undergoes the same processing operation as the low level feature fi(i=1,2,3,4) to obtain the feature Fi+1 with scale (64,hi+1,wi+1). In order to calculate interlayer attention, Fi+1 is upsampled (Up) and convolved to obtain Xa with scale (64,hi,wi), and then Xa performs element-wise multiplication with Fi to obtain preliminary attention Xa-p, which is the initial attention between high level and low level adjacent features with the scale of (64,hi,wi).In order to enhance the robustness of the ILA module, Xa-cat with the scale of (2×64,hi,wi) is obtained by concatenating the channel dimensions of Xa-p and Xa convolution results. After channel compression and Sigmoid activation, Xa-cat obtains the interlayer attention Ai(i=1,2,3,4) with a scale of (1,hi,wi).

The Ai calculation process is shown in (1)-(5):

$$X_i\text{-cat} = \text{cat}_c\big(X_i\text{-}0, X_i\text{-}1, X_i\text{-}2, X_i\text{-}3\big) \quad (1)$$

$$X_{i+1}\text{-cat} = \text{cat}_c\big(X_{i+1}\text{-}0, X_{i+1}\text{-}1, X_{i+1}\text{-}2, X_{i+1}\text{-}3\big) \quad (2)$$

$$Fi = \text{Relu}\big(\text{BN}\big(\text{Conv}_{464}\big(X_i\text{-cat}\big)\big)\oplus X_i\text{-}0\big) \quad (3)$$

$$Xa = \text{Conv}_{1\times1}\Big(\text{Up}\Big(\text{Relu}\Big(\text{BN}\Big(\text{Conv}_{464}\big(X_{i+1}\text{-cat}\big)\Big)\oplus X_{i+1}\text{-}0\Big)\Big)\Big) \quad (4)$$

$$Ai = \text{Sigmoid}\Big(\text{Conv}_{2641}\big(\text{cat}_c\big(Xa\otimes Fi, \text{Conv}_{1\times1}\big(Xa\big)\big)\big)\Big) \quad (5)$$

Where $\text{cat}_c(\cdot)$ represents latitude concatenation of channels; $\text{Conv}_{464}(\cdot)$ represents the convolution of input 4×64 channels and output 64 channels; $\oplus$ represents the addition of features and corresponding elements; $\text{BN}(\cdot)$ stands for normalization; $\text{Relu}(\cdot)$ indicates the Relu activation operation. $\text{Up}(\cdot)$ stands for upsampling; $\text{Conv}_{1\times1}(\cdot)$ stands for 1×1 convolution; $\otimes$ represents multiplication between feature map elements. $\text{Conv}_{2641}(\cdot)$ represents the convolution of input 2 × 64 channels and output 1 channels; $\text{Sigmoid}(\cdot)$ indicates the Sigmoid activation operation. In the following (6)-(16), $\text{cat}_c(\cdot)$, $\oplus$, $\text{BN}(\cdot)$, $\text{Relu}(\cdot)$, $\text{Up}(\cdot)$, $\otimes$, $\text{Sigmoid}(\cdot)$ have the same meaning as here and will not be explained later.

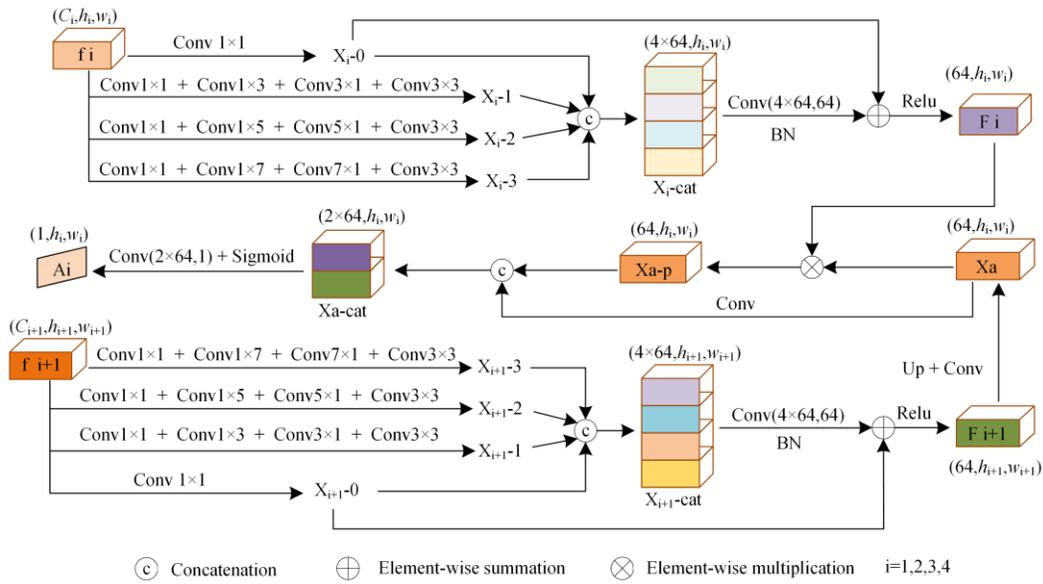

**Fig. 2.** Structure diagram of ILA module

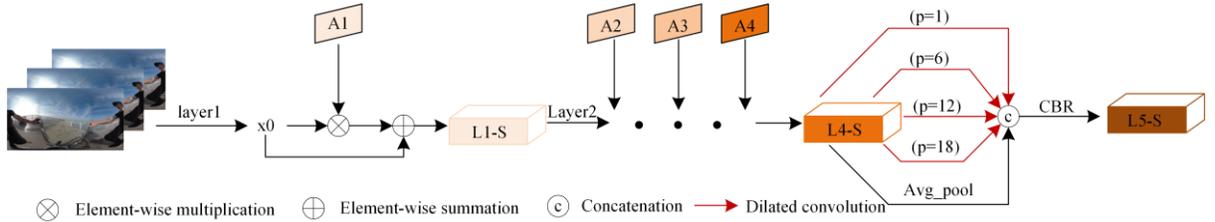

**Fig. 3.** Schematic diagram of the feature generation process of each layer of spatial flow

### 2.2.2. Feature generation of each layer

The output feature generation process of the five levels of spatial flow is shown in Fig. 3. Firstly, the layer1 of ResNet34 is used to extract the first level feature x0 of panoramic video frame, and x0 is multiplied with the corresponding elements of the interlayer attention A1, and then the multiplication result is added with the corresponding elements of x0 to obtain the first level feature L1-S of the spatial flow. Then, L1-S uses the layer2 of ResNet34 to extract features. The interlayer attention A2, A3, and A4 are applied to the corresponding level features in the same way to obtain the second, third, and fourth level features of the spatial flow, respectively. Then, the fourth level feature L4-S of spatial flow concatenates five kinds of results by ASPP operation, that is, through a parallel operation of dilated convolution (expansion rates is 1, 6, 12, and18) and average pooling. Finally, the fifth level feature L5-S of spatial flow is obtained by CBR operation.

## 2.3. Time flow processing

In the detection process of this paper, taking into account the relative motion between adjacent frames, the algorithm from literature [26] is used to calculate the optical flow of panoramic video, and the optical flow is used as one of the inputs to extract more accurate salient object information. In order to control complexity of the detection model, the classical and compact ResNet34 is selected to extract five levels salient object features contained in the optical flow. Optical flow comes from continuous frames of panoramic video, which contains relatively limited salient object information. ResNet34 network has been able to fully extract the salient object information of optical flow.

In addition, this paper also attempts to add depth map in the SOD at first, aiming at extracting more salient object features from depth information. However, after experiments, it is found that adding depth map cannot play a positive role in the SOD of panoramic video. The specific experimental results will be described in detail in the experimental section of this paper. There are two possible reasons why depth map cannot play a positive role in SOD of panoramic video. Firstly, the feature extraction method of panoramic video frames used in this paper is not suitable for feature extraction of depth map. Secondly, up to now, there is no published depth estimation algorithm for panoramic video, so this paper can only use the traditional 2D video depth estimation algorithm [27] to obtain the depth map of panoramic video, but the estimated depth map cannot accurately reflect the depth information of the panoramic video in fact.

## 2.4. Mixed flow processing

The function of the mixed flow processing is to integrate the features extracted from the spatial flow and the temporal flow, and then fuse the salient object information of the spatial-temporal dual flows mode to complete the SOD of panoramic video.

### 2.4.1. Interlayer weight module

The structure of the ILW module is shown in Fig. 4. The lowest level feature L1-S of the spatial flow is operated by CBR to obtain feature I1. L2-S extracts features through CBR operation and compresses the number of channels to 64, and then performs upsampling to the same spatial size as L1-S to obtain feature I2 (C-U represents CBR and upsampling operation in the figure, and C-U in other subsequent figures of this paper has the same meaning). Features I3, I4 and I5 are obtained in the same way as L2-S. The channel dimension of I1-I5 is concatenated to obtain the concatenated fusion feature cat-i with a scale of $(5 \times 64, h, w)$. The cat-i performs the convolution operation, compress the number of channels to 5, and then performs batch normalization（BN） and Relu activation operation to obtain the feature Interlayer-i with the scale of $(5, h, w)$ after feature fusion at each level of spatial flow. Interlayer-i with 5 channels reflects the salient object information contained in the features of each level of spatial flow. Interlayer-i is then pooled by global average, and the spatial flow weight-s with scale $(5, 1, 1)$ is obtained. The weight-s can reflect how much salient object information is contained in each level feature of spatial flow. The five level features of the temporal flow perform the same operations as the five level of the spatial flow to obtain the weight of the temporal flow weight-t. The weight-s and weight-t are concatenated in spatial dimensions, and then the Softmax operation is performed to obtain the Interlayer weight. The Interlayer weight represents the importance degree of salient object information contained in each layer feature and its scale is $(5, 2, 1)$.

The calculation process of Interlayer weight is shown in (6)-(8) :

$$\text{weight-s} = \text{Avgpool}\Big(\text{Relu}\big(\text{BN}\big(\text{Conv}_{564}\big(\text{cat}_c\left(\text{I1,I2,I3,I4,I5}\right)\big)\big)\big)\Big) \quad (6)$$

$$\text{weight-t} = \text{Avgpool}\Big(\text{Relu}\big(\text{BN}\big(\text{Conv}_{564}\big(\text{cat}_c\left(\text{F1,F2,F3,F4,F5}\right)\big)\big)\big)\Big) \quad (7)$$

$$\text{Interlayer weight} = \text{Softmax}\big(\text{cat}_s\left(\text{weight-s,weight-t}\right)\big) \quad (8)$$

Where  represents the convolution of 5×64 input channels and 5 output channels;  stands for global average pooling;  represents spatial latitude concatenation;  indicates the activation of Softmax. In the following (9)-(16),  have the same meaning as here and will not be explained later.

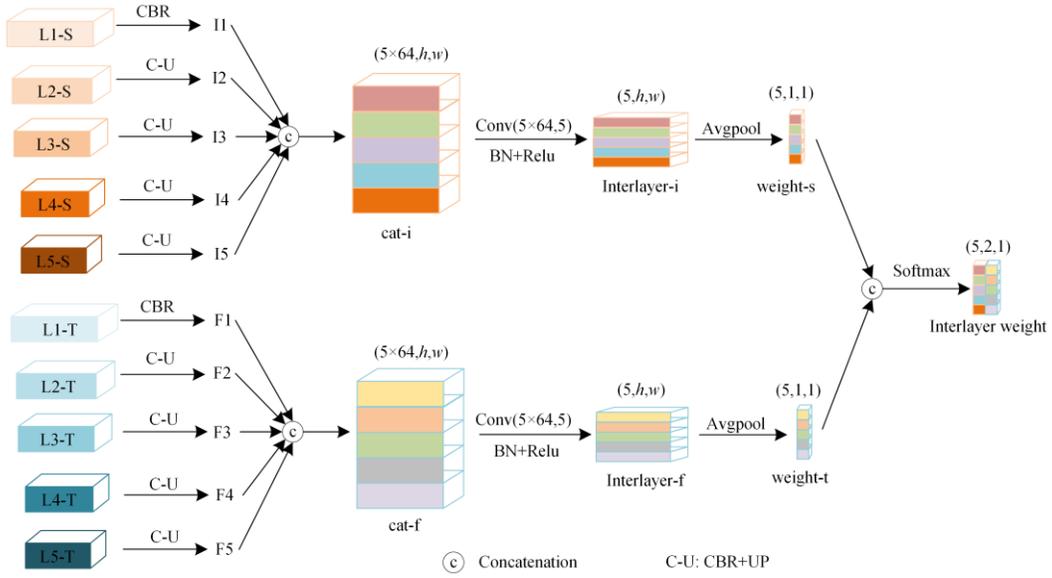

**Fig. 4.** Structure diagram of ILW module

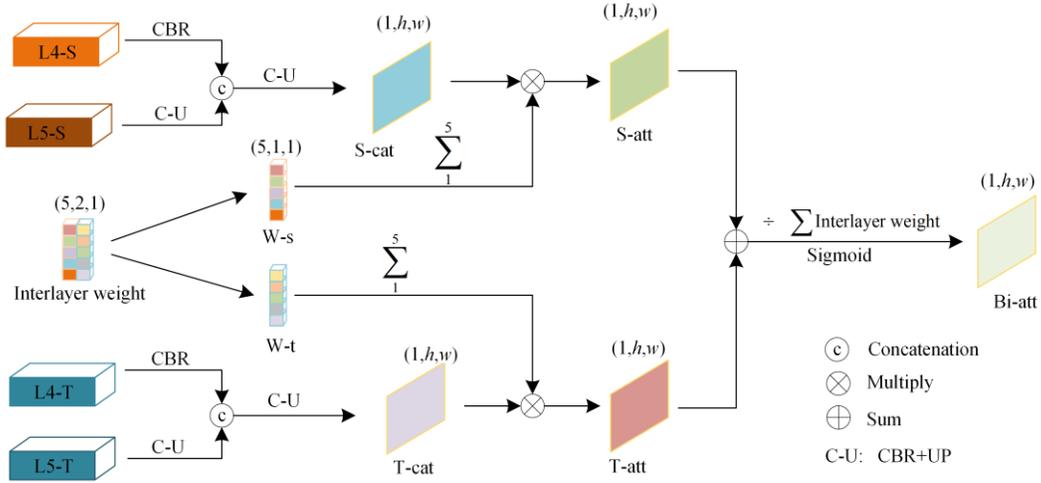

**Fig. 5.** Structure of bimodal attention

### 2.4.2. Bimodal attention module

The structure of the BMA module is shown in Fig. 5. CBR operation is performed on the high level feature L4-S of the spatial flow to extract features and compress channels, and L5-S is upsampled to the same spatial scale as L4-S after CBR operation, and the two results are concatenated. The concatenated results are extracted by CBR operation and the number of channels is compressed to 1. Then the results are upsampled to the lowest level feature spatial scale of spatial flow for mixed flow feature fusion, and the high-scale fusion feature S-cat of spatial flow with the scale of (1, h, w) is obtained. The Interlayer weight is divided into W-s and W-t with the scale of (5, 1, 1) from the spatial dimension, which represent the interlayer weight of spatial flow and temporal flow respectively. The sum of W-s is taken as the weight of the spatial flow fusion feature S-cat, S-cat and W-s are multiplied to obtain the spatial flow attention S-att with the scale of (1, h, w). The temporal flow high level features are processed in the same way as the spatial flow high level features to obtain the temporal flow attention T-att. S-att and

T-att are added and then divided by the sum of Interlayer weight to achieve normalization. Finally, Sigmoid activation is performed to obtain a bimodal attention Bi-att with a scale of (1, h, w).

The relationship between high-level features, interlayer weights and Bi-att is shown in (9)-(11) :

$$\text{S-att} = \text{CU}\Big(\text{cat}_c\big(\text{CBR}(\text{L4-S}), \text{CU}(\text{L5-S})\big)\Big) \otimes \sum(\text{W-s}) \quad (9)$$

$$\text{T-att} = \text{CU}\Big(\text{cat}_c\big(\text{CBR}(\text{L4-T}), \text{CU}(\text{L5-T})\big)\Big) \otimes \sum(\text{W-t}) \quad (10)$$

$$\text{Bi-att} = \text{Sigmoid}\Big(\big(\text{S-att} \oplus \text{S-att}\big) \div \sum(\text{Interlayer weight})\Big) \quad (11)$$

Where represents convolution, normalization and modified linear unit activation operations; represents and upsampling operations; stands for sum. In the following (12)-(16), has the same meaning as here and will not be explained later.

### 2.4.3. Multi-level feature fusion

The multi-level feature fusion structure is shown in Fig. 6. The interlayer weight is composed of the spatial flow interlayer weight Swi and the temporal flow Interlayer weight

Twi (i=1,2,3,4,5). Inspired by the literature [23], new interlayer weights are obtained as below:

$$\begin{cases} \text{Twi} = 0 \text{, if Swi-Twi} \geq 0.5 \\ \text{Swi} = 0 \text{, if Swi-Twi} \leq -0.5 \\ \text{Swi} = \text{Swi, Twi} = \text{Twi , others} \end{cases} \quad (12)$$

That is, if the absolute value of the difference between Swi and Twi exceeds 0.5, the smaller weight value is assigned to 0.The threshold processed spatial flow interlayer weight Swi and temporal flow interlayer weight Twi are respectively multiplied with the corresponding hierarchical features to filter out non-salient object features, and then the hierarchical features at the same level are added together to obtain Mix1-Mix5 through CBR operation. Upsampling Mix5 with scale $(c, h_5, w_5)$ yields Fup5 with scale $(c, h_4, w_4)$. Fup5 is added with Mix 4 of the same scale, and the results are fused by CBR operation for feature fusion and then Fup4 of scales$(c, h_3, w_3)$ is obtained by upsampling.Fup4 undergoes layer-by-layer feature fusion and upsampling by the same method, and then Fup3 and Fup2 are obtained in turn. Mix 1 and the fusion feature Fup2 are added and then CBR operation is performed to obtain Fup1 with scale $(64, h_1, w_1)$. Fup1 contains the most abundant salient object information, and then it is multiplied with the bimodal attention Bi-att to further strengthen the salient object information and obtain Fa with the scale of $(64, h_1, w_1)$. Fa and Fup1 are added to ensure module performance. The sum result is first operated by CBR, then the number of channels is compressed by convolution, and finally it is upsampled to the spatial scale of the input video to get the final SOD result OUT.

Fup4 can be calculated by (13)-(15) as:

$$\text{Fup5} = \text{Up}\big(\text{CBR}\big((\text{L5-S}\otimes\text{iw}_{s5})\oplus(\text{L5-T}\otimes\text{iw}_{t5})\big)\big) \quad (13)$$

$$\text{Mix4} = \text{CBR}\big((\text{L4-S}\otimes\text{iw}_{s4})\oplus(\text{L4-T}\otimes\text{iw}_{t4})\big) \quad (14)$$

$$\text{Fup4} = \text{Up}\big(\text{CBR}\big(\text{Fup5}\oplus\text{Mix4}\big)\big) \quad (15)$$

Where $\text{iw}_{si}$ and $\text{iw}_{ti}$ (i=4, 5) represent the weight of corresponding layers of space flow and time flow in Interlayer weight respectively. The formulas for Fup1, Fup2, and Fup3 are similar to those for Fup4 and are no longer listed. The output OUT of multi-level feature fusion can be expressed by (16):

$$\text{OUT} = \text{Up}\big(\text{Conv}_{641}\big(\text{CBR}\big(\text{Bi-att}\otimes\text{Fup1}\oplus\text{Fup1}\big)\big)\big) \quad (16)$$

Where $\text{Conv}_{641}(\cdot)$ represents the convolution of 64 input channels and 1 output channels;

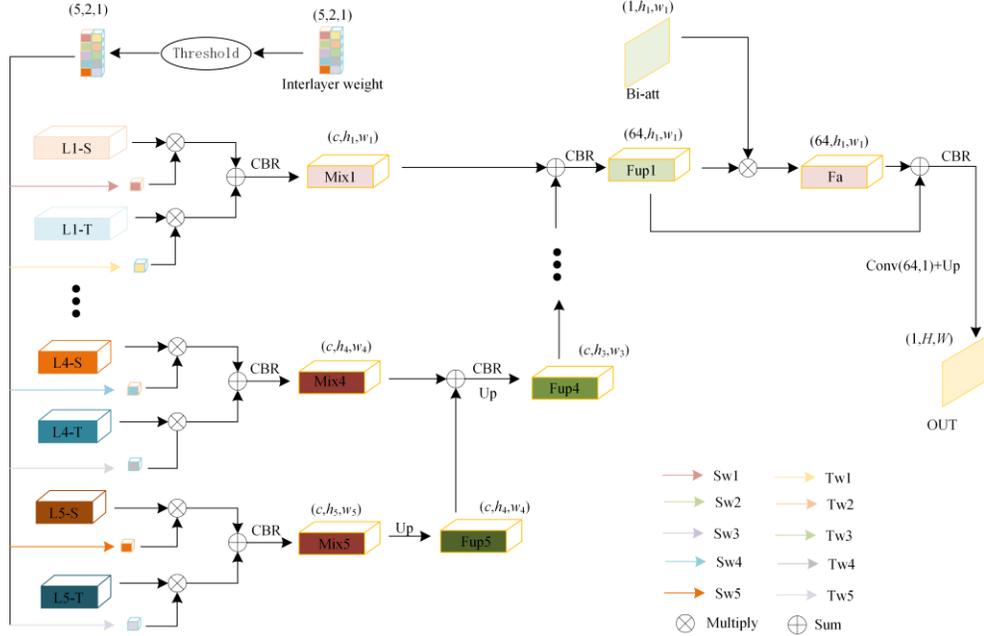

**Fig. 6.** Structure diagram of multi-level feature fusion

### 2.5. Multiple mixed flow loss function

In order to make the proposed STDMMF-Net converge quickly, this paper adopts multi-level supervision for training. The loss function consists of three parts, namely spatial flow, temporal flow, and mixed flow loss function.

The spatial flow loss function takes the highest level feature L5-S of the spatial flow as the supervision target. The specific spatial flow loss function can be expressed as follows:

$$\text{loss1} = \text{loss}_{\text{BCE}}\big(I_{\text{sal}}, G\big) \quad (17)$$

Where $I_{\text{sal}}$ represents the feature map of L5-S after convolution and upsampling operation, G represents the ground truth map, and $\text{loss}_{\text{BCE}}$ represents the Binary Cross Entropy (BCE) loss function.

The time flow loss function takes the highest level feature L5-T of the time flow as the supervision target, and the specific time flow loss function can be expressed as follows:

$$\text{loss2} = \text{loss}_{\text{BCE}}\big(F_{\text{sal}}, G\big) \quad (18)$$

Here, $F_{\text{sal}}$ represents the feature map of L5-T after convolution and upsampling operation.

The mixed flow loss function takes the final output of the model as the supervision target and can be expressed as:

$$\text{loss3} = \text{loss}_{\text{BCE}}\left(\text{OUT}, G\right) \quad (19)$$

Here, OUT is the salient object map of the final output of the proposed model and G is the ground truth map.

In this paper, it is found that the overall loss function has the best performance when the weight value of the spatial flow loss function and the time flow loss function are 0.6 and 0.4, respectively, that is

$$\text{loss} = 0.6\text{loss1} + 0.4\text{loss2} + \text{loss3} \quad (20)$$

## 3. EXPERIMENTAL RESULTS AND ANALYSIS

### 3.1. Data set and experiment setup

In this paper, we use two panoramic video datasets SHD360[8] and ASOD60K[9] to carry out experiments. SHD360 contains 41 panoramic videos with a total of 6268 frames and a total duration of 1119s. ASOD60K contains 68 panoramic videos with a total of 10300 frames and a total duration of 1922s.

We use Pytorch architecture and 65 epochs to train the proposed STDMMF-Net, and four consecutive frames of panoramic video are processed in batches as a group. The learning rate is set to 0.0001, and the SGD optimizer with momentum parameter of 0.9 is adopted, and weight decay of 0.00001 is used. All experiments in this paper were done on a desktop computer equipped with an RTX3060 GPU and an AMD I5 3600 CPU.

### 3.2. Experimental comparison before and after adding depth flow

In this section, we try to add depth flow to the detection, that is, the depth map is obtained by depth estimation for each frame of the panoramic video, and more salient object features are extracted from the depth information to test the role of depth map in the panoramic video SOD task. Three types of comparative experiments were carried out.

The first type: The deep flow performs the same feature extraction operation as the optical flow. Three different thresholds (0, 0.2, 0.4) are used to fuse the multi-level features in the mixed flow.

The second type: after the deep flow extracts features through ResNet34, the interlayer attention is obtained from the depth features of each layer, and the interlayer attention assists the spatial flow to extract features to obtain the features of each level of the spatial flow.

The third type: using depth map as loss weight to supervise model training.

The experimental comparison results of the above three types of attempted methods and the proposed STDMMF-Net on the datasets SHD360 and ASOD60K are shown in Table I and Table II, respectively. In this paper, we use the Mean Absolute Error (MAE), max-F, mean-F, max-Em, mean-Em, and Structure-measure (Sm) to compare the performance of different methods. From the experimental results, it can be seen that for the six evaluation indexes, the detection results of the three types of methods tried on the two data sets are inferior to the proposed method. Therefore, this paper finally adopts two modes of spatial flow and temporal flow to detect salient objects in panoramic video.

Table I
Experimental comparison before and after adding depth flow on the SHD360 dataset

| Methods | | MAE↓ | max-F↑ | mean-F↑ | max-Em↑ | mean-Em↑ | Sm↑ |
|---|---|---|---|---|---|---|---|
| Type | Threshold | | | | | | |
| 1 | 0 | 0.0348 | 0.7249 | 0.6949 | 0.8981 | 0.8596 | 0.8078 |
| | 0.2 | 0.0360 | 0.7206 | 0.6928 | 0.8924 | 0.8476 | 0.7998 |
| | 0.4 | 0.0371 | 0.6779 | 0.6524 | 0.8630 | 0.8157 | 0.7703 |
| 2 | / | 0.0346 | 0.7204 | 0.6935 | 0.8910 | 0.8587 | 0.8122 |
| 3 | / | 0.0339 | 0.7340 | 0.7066 | 0.9017 | 0.8664 | 0.8134 |
| Ours | / | **0.0323** | **0.7444** | **0.7153** | **0.9083** | **0.8773** | **0.8179** |

Table II
Experimental comparison before and after adding depth flow on the ASOD60K dataset

| Methods | | MAE↓ | max-F↑ | mean-F↑ | max-Em↑ | mean-Em↑ | Sm↑ |
|---|---|---|---|---|---|---|---|
| Type | Threshold | | | | | | |
| 1 | 0 | 0.0353 | 0.1908 | 0.1700 | 0.5873 | 0.5486 | 0.5741 |
| | 0.2 | 0.0329 | 0.2128 | 0.1961 | 0.5896 | 0.5642 | 0.5919 |
| | 0.4 | 0.0366 | 0.1847 | 0.1510 | 0.6069 | 0.5357 | 0.5621 |
| 2 | / | 0.0326 | 0.2016 | 0.1843 | 0.5923 | 0.5583 | 0.5834 |
| 3 | / | 0.0319 | 0.1967 | 0.1696 | 0.6082 | 0.5451 | 0.5730 |
| Ours | / | **0.0290** | **0.3287** | **0.3156** | **0.6671** | **0.6490** | **0.6371** |

### 3.3. Experimental comparison on different datasets

In this paper, the proposed model is compared with seven advanced indirect panoramic video SOD methods, including BBSNet[24], MQP[21], DFMNet[22], ALGRF[20], DCFNet[18], DCTNet[19] and DSNet[23]. For a fair comparison, different comparison methods are tested in the same experimental environment. Since the panoramic video SOD method based on audio and video information proposed in the literature [10] has limited application scenarios, this paper does not compare with it.

#### 3.3.1. Experimental comparison on the SHD360 dataset

Table III shows the quantitative comparison results of the six evaluation metrics on the SHD360 dataset. It can be seen from the experimental results that performance of the proposed model exceeds that of all comparison methods under the six evaluation metrics. Specifically, compared with the best indexes of all the excellent algorithms compared(marked in red in Table III), the optimization ranges of the proposed model are 2.17%, 1.52%, 0.70%, 1.01%, 0.46% and 0 respectively in MAE, max-F, mean-F, max-Em, mean-Em and Sm indexes.

Table III
Comparison of objective metrics of each model on the SHD360 dataset

| Methods | MAE↓ | max-F↑ | mean-F↑ | max-Em↑ | mean-Em↑ | Sm↑ |
|---------|------|--------|---------|---------|----------|-----|
| Ours | **0.0323** | **0.7444** | **0.7153** | **0.9083** | **0.8773** | **0.8179** |
| BBSNet | 0.0848 | 0.2958 | 0.2281 | 0.7508 | 0.5892 | 0.5668 |
| MQP | 0.0440 | 0.6610 | 0.6361 | 0.8526 | 0.8289 | 0.7755 |
| DFMNet | 0.0594 | 0.4044 | 0.3237 | 0.7763 | 0.6296 | 0.6010 |
| ALGRF | 0.0421 | 0.6526 | 0.6244 | 0.8292 | 0.8136 | 0.7659 |
| DCFNet | 0.1062 | 0.1180 | 0.1116 | 0.5184 | 0.5098 | 0.4676 |
| DCTNet | 0.0619 | 0.4235 | 0.3848 | 0.7850 | 0.7005 | 0.6160 |
| DSNet | 0.0330 | 0.7336 | 0.7103 | 0.8992 | 0.8733 | 0.8117 |

To further verify the performance of the proposed model, a visual comparison is performed with seven SOTA methods. The results are shown in Fig. 7. The results show that the proposed method can more accurately detect the salient objects in panoramic videos. For example, Fig. 7 (a) and (b) show that the proposed method can more accurately detect salient information such as the leg of the person. Fig. 7 (c) and (d) show that the salient objects detected by the proposed method are more complete, and the boundaries are closer to the true values of manual labeling. Fig. 7 (e) and (f) show that the proposed method has a better detection effect for small salient objects.

The objective metrics in Table III and the visual comparison results in Fig. 7 prove that the performance of the proposed model on the SHD360 dataset is better than that of the existing representative advanced methods.

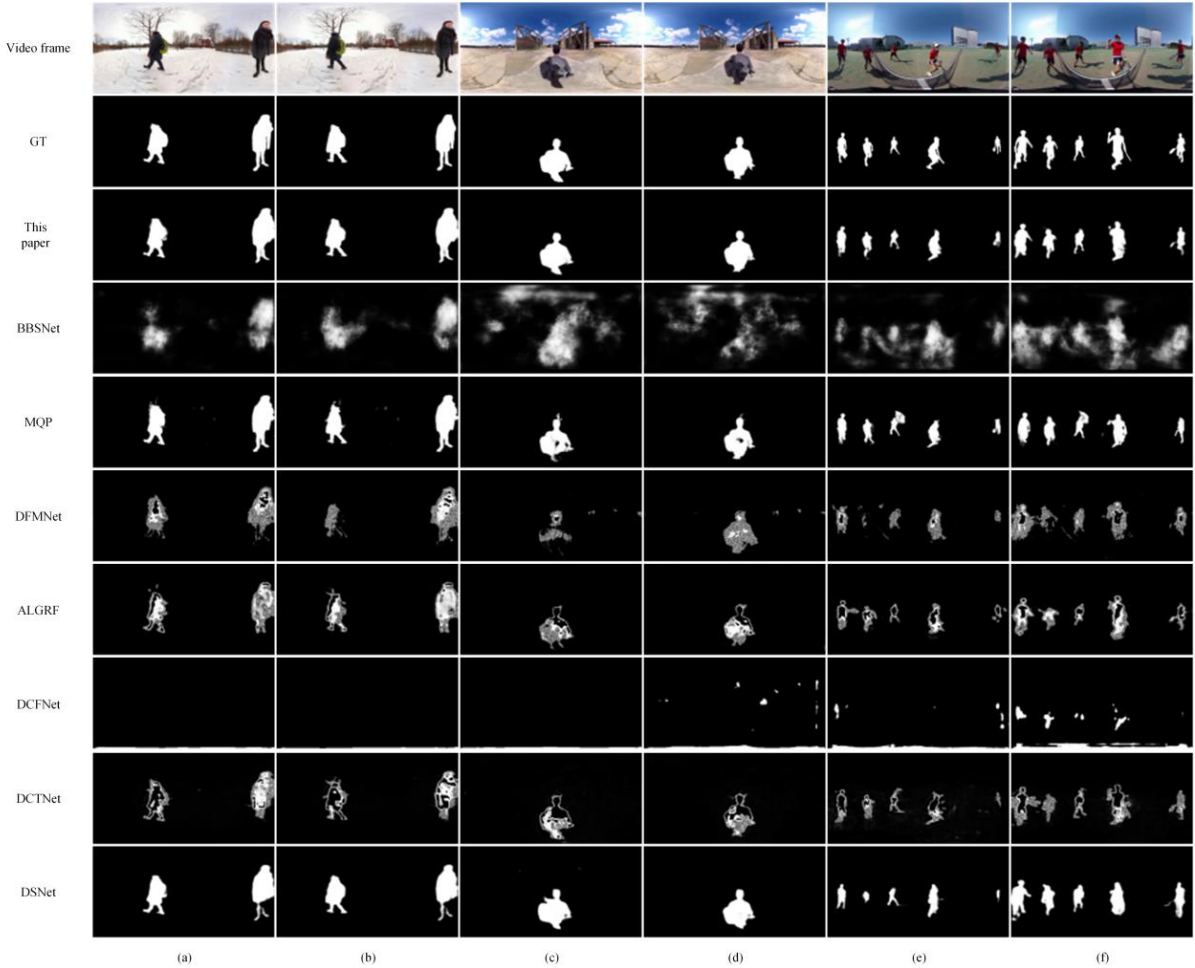

**Fig. 7.** Visual comparison results of the proposed model and the listed SOTA methods on the SHD360 dataset

### 3.3.2. Experimental comparison on the ASOD60K dataset

Table IV shows the quantitative comparison results of the six evaluation metrics on the ASOD60K dataset. It can be seen from the experimental results that the performance of the proposed model exceeds the performance of all comparison methods under the six evaluation metrics. Specifically, in terms of MAE, max-F, mean-F, max-Em, mean-Em, and Sm metrics, compared with the best indexes of all the excellent algorithms compared (marked in red in Table IV), the optimization ranges of the proposed model are 22.41%, 2.49%, 4.30%, 2.35%, 3.48% and 0.52%, respectively.

Table IV

Comparison of objective metrics of each model on the
ASOD60K dataset

| Methods | MAE↓ | max-F↑ | mean-F↑ | max-Em↑ | mean-Em↑ | Sm↑ |
|---------|------|--------|---------|---------|----------|-----|
| Ours | **0.0290** | **0.3287** | **0.3156** | **0.6671** | **0.6490** | **0.6371** |
| BBSNet | 0.0819 | 0.0816 | 0.0634 | 0.5731 | 0.4808 | 0.4995 |
| MQP | 0.0447 | 0.2782 | 0.2516 | 0.6346 | 0.5756 | 0.6071 |
| DFMNet | 0.0896 | 0.1545 | 0.1256 | 0.6095 | 0.5323 | 0.5223 |
| ALGRF | 0.0440 | 0.3043 | 0.2709 | 0.6305 | 0.5640 | 0.6037 |
| DCFNet | 0.2798 | 0.0557 | 0.0540 | 0.3762 | 0.3722 | 0.3887 |
| DCTNet | 0.0577 | 0.1706 | 0.1561 | 0.6047 | 0.5559 | 0.5395 |
| DSNet | 0.0355 | 0.3207 | 0.3026 | 0.6518 | 0.6272 | 0.6338 |

To further verify the performance of the proposed model, a visual comparison with seven SOTA methods is performed, as shown in Fig. 8. The comparison results show that the proposed method can detect salient objects in panoramic videos more accurately. For example, as shown in Fig. 8 (a) and (b), the proposed method can better filter out background interference and accurately detect details such as the legs of characters. In Fig. 8 (c) and (d), the proposed method can detect small salient objects with similar color to the background more accurately. In Fig. 8 (e) and (f), the proposed method can resist strong interference signals such as tables and chairs, and detect salient objects more accurately and completely.

The objective metrics in Table IV and the visual comparison results in Fig. 8 prove that the performance of the proposed model on the ASOD60K dataset is better than that of the existing representative advanced methods.

It should be noted that the detection metric of the proposed method on the SHD360 dataset is generally better than that on the ASOD60K dataset. There are two reasons for this result. First, ASOD60K data set contains more full-motion video and has more complex scenes, which significantly increases the difficulty of SOD detection on it; Second, ASOD60K is a panoramic video SOD dataset based on audio guidance. In order to compare the proposed method fairly with more general SOTA methods, this paper conducts SOD detection without using audio signals, which increases the difficulty and inevitably reduces the detection effect.

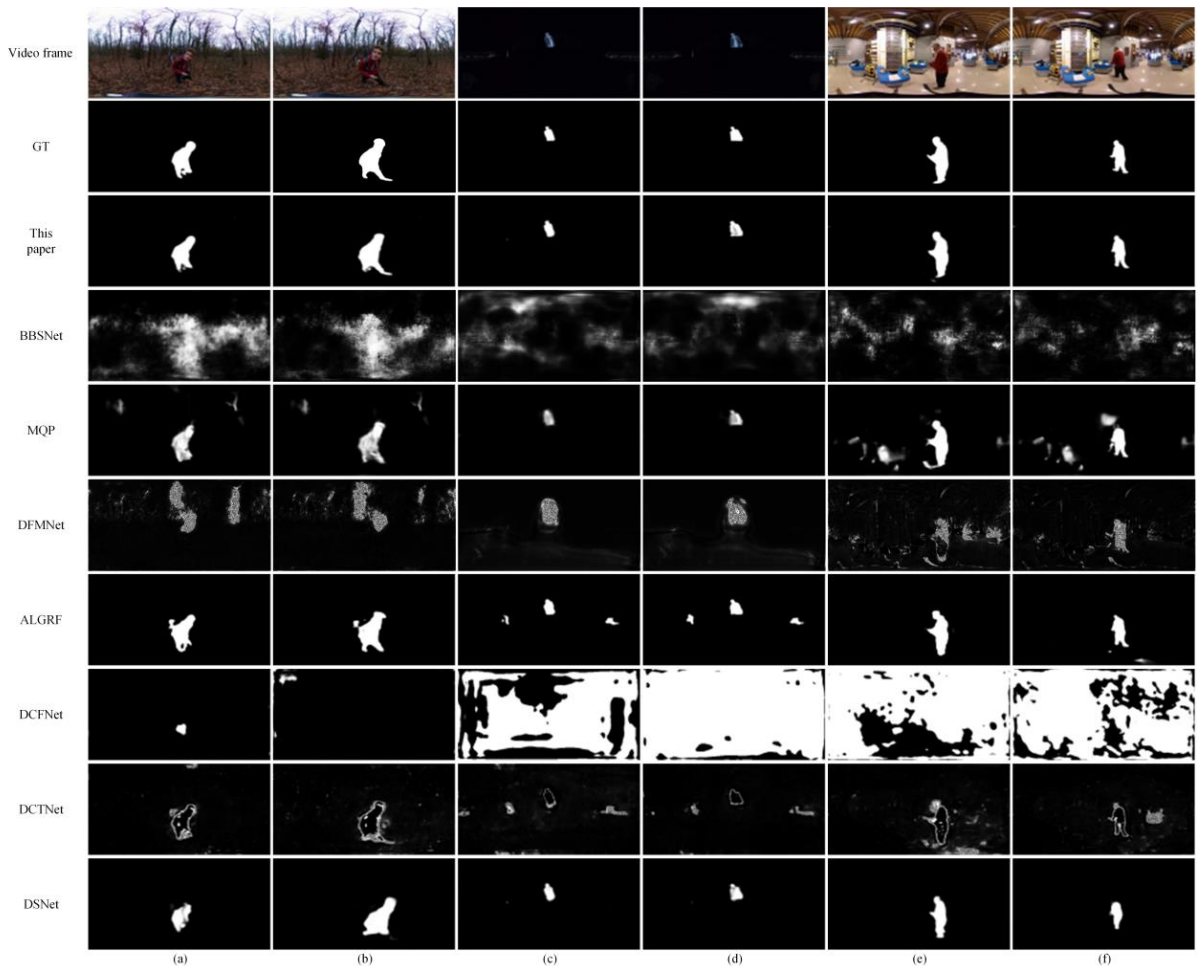

**Fig. 8.** Visual comparison results of the proposed model and the listed SOTA methods on the ASOD60K dataset

### 3.4. Ablation Experiments

#### 3.4.1. Modal ablation experiments

In order to verify the effectiveness of the dual-mode method including video frame and optical flow for SOD in panoramic video, ablation experiments were carried out. The experimental results are shown in Table V. The spatial flow in the table represents only using video frames as model inputs to extract salient information of panoramic video from spatial dimension. The time flow represents only using optical flow as model input to extract salient information of panoramic video from time dimension. Mixed flow represents the proposed model, that is, both video frames and optical flows are used as model inputs. As can be seen from the experimental results in Table V, the experimental results of spatial flow are better than that of time flow in all six metrics, which is due to the fact that video frames contain more salient object information of panoramic video than optical flow. The experimental results of mixed flow are better than those of spatial flow, and the optimization amplitude in MAE, max-F, mean-F, max-Em, mean-Em, and Sm metrics is 15.17%, 13.25%, 12.31%, 1.71%, 2.09% and 4.24%, respectively. The above comparison results show that the mixed flow proposed in this paper can significantly improve the performance metrics and enhance the SOD ability of panoramic video.

Table V
Ablation results of different modes

| Mode | MAE↓ | max-F↑ | mean-F↑ | max-Em↑ | mean-Em↑ | Sm↑ |
|---|---|---|---|---|---|---|
| spatial flow | 0.0372 | 0.6573 | 0.6369 | 0.8930 | 0.8593 | 0.7846 |
| time flow | 0.0531 | 0.4556 | 0.4409 | 0.7470 | 0.6922 | 0.6448 |
| mixed flow | **0.0323** | **0.7444** | **0.7153** | **0.9083** | **0.8773** | **0.8179** |

#### 3.4.2. Module ablation experiments

To verify the effectiveness of relevant modules of the proposed model, ablation experiments were conducted. The baseline represents ResNet34, and then gradually adds ILA module, ILW module, and BMA module. The experimental results are shown in Table VI. The comparison shows that the six metrics are all improved by adding ILA module on the basis of Baseline. Further adding ILW module, except mean-F metric decreased slightly, the other five metrics increased. The reason for the decrease of mean-F index may be that the interlayer weights reduce the salient object feature regions at each level of mixed flow, which leads to the decrease of regional similarity. Finally, BMA module is added to form the complete model of this paper, and all the metrics are improved. Specifically, compared with Baseline, the optimization ranges of MAE, max-F, mean-F, max-Em, mean-Em and Sm are 81.14%, 4.02%, 16.16%, 11.80%, 12.40% and 9.54% respectively.

The above comparison results show that the ILA module, ILW module, and BMA module proposed in this paper can enhance the ability of SOD in panoramic video, and are effective for obtaining better detection results.

Table VI
Results of ablation experiments for related modules

| Baseline | ILA | ILW | BMA | MAE↓ | max-F↑ | mean-F↑ | max-Em↑ | mean-Em↑ | Sm↑ |
|---|---|---|---|---|---|---|---|---|---|
| √ | | | | 0.0598 | 0.7156 | 0.6158 | 0.8124 | 0.7805 | 0.7467 |
| √ | √ | | | 0.0430 | 0.7246 | 0.6689 | 0.8159 | 0.8472 | 0.7941 |
| √ | √ | √ | | 0.0359 | 0.7260 | 0.6671 | 0.8897 | 0.8579 | 0.8127 |
| √ | √ | √ | √ | **0.0323** | **0.7444** | **0.7153** | **0.9083** | **0.8773** | **0.8179** |

### 3.5. Experimental comparison of generalization performance

To verify the generalization performance of the proposed model, a comparison experiment with the existing SOTA models is carried out. The model is trained on the SHD360 dataset and tested on the ASOD60K dataset, and the experimental results are shown in Table VII. The best metrics of the comparison methods are marked in red. Except for the max-Em metric, the optimization ranges of the proposed model in MAE, max-F, mean-F, mean-Em, and Sm are 13.48%, 13.20%, 18.93%, 11.66%, and 2.76%, respectively, compared with the best metric of the SOTA models in Table VII. The above data show that the proposed model has better generalization performance than the SOTA models.

Table VII
Comparison of generalization performance of different models

| Methods | MAE↓ | max-F↑ | mean-F↑ | max-Em↑ | mean-Em↑ | Sm↑ |
|---|---|---|---|---|---|---|
| Ours | **0.0282** | **0.1973** | **0.1866** | 0.5927 | **0.5696** | **0.5842** |
| BBSNet | 0.1091 | 0.0565 | 0.0431 | 0.6194 | 0.4109 | 0.4800 |
| MQP | 0.0436 | 0.1701 | 0.1569 | 0.5467 | 0.5101 | 0.5685 |
| DFMNet | 0.1749 | 0.0866 | 0.0702 | 0.5482 | 0.4662 | 0.4788 |
| ALGRF | 0.0320 | 0.1743 | 0.1529 | 0.5628 | 0.5002 | 0.5626 |
| DCFNet | 0.0619 | 0.0390 | 0.0386 | 0.5109 | 0.4962 | 0.4879 |
| DCTNet | 0.0461 | 0.0740 | 0.0694 | 0.6544 | 0.5077 | 0.5132 |

### 3.6. Comparison experiments of memory required for model inference and test time

To measure the performance of different methods more comprehensively, this paper compares the memory required for model inference and the time required for model testing with other SOTA models. Since the test time of different models on any data set has the same distribution pattern and due to space limitations, this paper only provides the test time of different models on the ASOD60K data set. The comparison results are shown in Table VIII. Compared with BBSNet, DFMNet, ALGRF, DCFNet, and DCTNet, the reasoning memory required by the proposed method is reduced by 205.73%, 1.81%, 56.03%, 44.22%, and 5.85% respectively. The test time is reduced by 1082.49%, 64.10%, 253.86%, 208.35%, and 416.20%, respectively. Compared with DSNet, the proposed model requires 33.88% (161.88MB) more memory for inference but reduces the test time by 233.84% (137.71 seconds). The above experimental results show that compared with other SOTA models, the proposed method can better control the memory required for model inference under the premise of using the shortest test time.

Table VIII
Comparison of memory required for inference and test time for different models

| Memory required for inference/ test time | Ours | BBSNet | DFMNet | ALGRF | DCFNet | DCTNet | DSNet |
|---|---|---|---|---|---|---|---|
| Memory(MB) | 477.83 | 1460.88 | 486.46 | 745.54 | 689.12 | 505.78 | **315.98** |
| Time(s) | **58.89** | 723.37 | 96.64 | 208.39 | 181.59 | 303.99 | 196.60 |

### 3.7. Complexity comparison

The execution time of the network will directly affect the efficiency of SOD in panoramic video. Therefore, this paper uses the metric Floating Point of Operations (FLOPs) to represent the model complexity, and compares the complexity of different models. The comparison results are shown in Table IX. Compared with MQP, ALGRF, DCFNet, and DCTNet, the computational complexity of the proposed method is reduced by 191.39%, 102.36%, 64.76%, and 42.79%, respectively. The above results show that the complexity of the proposed model is well-controlled.

Table IX
Comparison of the complexity of different models

| Complexity | Ours | BBSNet | MQP | DFMNet | ALGRF | DCFNet | DCTNet | DSNet |
|---|---|---|---|---|---|---|---|---|
| FLOPs(G) | 86.3078 | 59.0256 | 251.4935 | **22.2249** | 174.6550 | 142.2016 | 123.2372 | 59.9480 |

## 4. CONCLUSION

This paper proposed STDMMF-Net, which uses spatial flow and optical flow for SOD in panoramic video. The model consists of three modules: the ILA module is used to improve the accuracy of salient object feature detection in spatial flow. The ILW module is used to quantify the salient object information contained in the features of each layer of the mixed stream, so as to improve the fusion efficiency of salient object features at each level. The BMA module is used to improve the SOD accuracy of the model. Extensive experiments on two datasets demonstrate that the proposed model has better accuracy in SOD of panoramic video than the existing SOTA methods. At the same time, the overall performance of the proposed model is better in terms of memory required for model inference, testing time, complexity, and generalization performance. In the future, we will carry out the following work: exploring the depth estimation algorithm for panoramic video and adding depth flow to SOD method; adding more modes as detection input to improve the efficiency of SOD in panoramic video.

## REFERENCES

[1] A. Kompella, R. V. Kulkarni, "A semi-supervised recurrent neural network for video salient object detection," Neural Computing and Applications, vol. 33, no. 6, pp. 2065-2083, Mar. 2021, doi: https://doi.org/10.1007/s00521-020-05081-5.

[2] G. Ma, S. Li, C. Chen, A. Hao, and H. Qin , "Stage-wise salient object detection in 360 omnidirectional images via object-level semantical saliency ranking," IEEE Trans. Visualization and Computer Graphics, vol. 26, no. 12, pp. 3535-3545, Dec. 2020, doi: https://doi.org/10.1007/s00521-020-05081-5.

[3] C. Chen, S. Li, Y. Wang, H. Qin, and A. Hao, "Video saliency detection via spatial-temporal fusion and low-rank coherency diffusion," IEEE Trans. Image processing, vol. 26, no.7, pp. 3156-3170, Feb. 2017, doi: 10.1109/TIP.2017.2670143.

[4] W. Wang, J. Shen, J. Xie, M. M. Cheng, H. Ling, and A. Borji, "Revisiting video saliency prediction in the deep learning era," IEEE Trans. Pattern Analysis and Machine Intelligence, vol. 43, no. 1, pp. 220-237, Jun. 2019, doi: 10.1109/TPAMI.2019.2924417.

[5] W. Wang, Q. Lai, H. Fu, J. Shen, H. Ling, and R. Yang, "Salient object detection in the deep learning era: An in-depth survey," EEE Trans. Pattern Analysis and Machine Intelligence, vol. 44, no. 6, pp. 3239-3259, Jan. 2021, doi: 10.1109/TPAMI.2021.3051099.

[6] H. B. Bi, D Lu, H. H. Zhu, L. N. Yang, and H. P. Guan, "STA-Net: spatial-temporal attention network for video salient object detection," Applied Intelligence, vol. 51, pp. 3450-3459, Sept. 2021, doi: https://doi.org/10.1007/s10489-020-01961-4.

[7] C. Chen, H. Wang, Y. Fang, and C. Peng, "A novel long-term iterative mining scheme for video salient object detection," IEEE Trans. Circuits and Systems for Video Technology, vol. 32, no. 11, pp. 7662-7676, Jun. 2022, doi: 10.1109/TCSVT.2022.3185252.

[8] Y. Zhang, L. Zhang, K. Wang, W. Hamidouche, and O. Deforges, "SHD360: A Benchmark Dataset for Salient Human Detection in 360° Videos," arXiv e-prints, May. 2021, doi: 10.48550/arXiv.2105.11578.

[9] Y. Zhang, "ASOD60K: An Audio-Induced Salient Object Detection Dataset for Panoramic Videos," arXiv, Nov. 2021, doi: https://doi.org/10.48550/arXiv.2107.11629.

[10] X. Li, H. Cao, S. Zhao, J. Li, L. Zhang, and B. Raj, "Panoramic Video Salient Object Detection with Ambisonic Audio Guidance," arXiv, Nov. 2022, doi: https://doi.org/10.48550/arXiv.2211.14419.

[11] W. Wang, J. Shen, L. Shao, "Video salient object detection via fully convolutional networks," IEEE Trans. Image Processing, vol. 27, no. 1, pp.38-49, Sept. 2017, doi: 10.1109/TIP.2017.2754941.

[12] H. Song, W. Wang, S. Zhao, J. Shen, and K. M. Lam, "Pyramid dilated deeper convlstm for video salient object detection." in ECCV, 2018, pp. 715-731.

[13] M. Shokri, A. Harati, K. Taba, "Salient object detection in video using deep non-local neural networks," journal of Visual Communication and Image Representation, vol. 68, pp.102769, 2020, doi: https://doi.org/10.1016/j.jvcir.2020.102769.

[14] X. Zhao et al, "Motion-aware Memory Network for Fast Video Salient Object Detection," arXiv, Aug. 2022, doi: https://doi.org/10.48550/arXiv.2208.00946.

[15] B. E. Bejnordi, A. Habibian, F. Porikli, and A. Ghodrati, "SALISA: Saliency-Based Input Sampling for Efficient Video Object Detection," arXiv, Apr. 2022, doi: https://doi.org/10.48550/arXiv.2204.02397.

[16] Y. Jiao et al, "Guidance and teaching network for video salient object detection," ICIP, Anchorage, AK, USA, 2021, pp. 2199-2203.

[17] Y. Su, J. Deng, R. Sun, G. Lin, H. Su, and Q. Wu, "A unified transformer framework for group-based segmentation: Co-segmentation, co-saliency detection, and video salient object detection," IEEE Trans. Multimedia, Apr. 2023, doi: 10.1109/TMM.2023.3264883.

[18] M. Zhang et al, "Dynamic context-sensitive filtering network for video salient object detection," ICCV, 2021, pp. 1553-1563.

[19] Y. Lu, D. Min, K. Fu, and Q. Zhao, "Depth-cooperated trimodal network for video salient object detection," ICIP, 2022, pp.116-120.

[20] Y. Tang, Y. Li, G. Xing, "Video salient object detection via adaptive local-global refinement," arXiv, 2021, doi: https://doi.org/10.48550/arXiv.2104.14360.

[21] C. Chen, J. Song, C. Peng, G. Wang, and Y. Fang, "A novel video salient object detection method via semisupervised motion quality perception," IEEE Trans. Circuits and Systems for Video Technology, vol. 32, no. 5, pp.2732-2745, May. 2022, doi: 10.1109/TCSVT.2021.3095843.

[22] W. Zhang, G. P. Ji, Z. Wang, K. Fu, and Q. Zhao, "Depth quality-inspired feature manipulation for efficient RGB-D salient object detection," Proceedings of the 29th ACM international conference on multimedia, 2021, pp.731-740.

[23] J. Liu, J. Wang, W. Wang and Y. Su, "DS-Net: Dynamic spatiotemporal network for video salient object detection," Digital Signal Processing, vol. 130, pp. 103700, Sept. 2022, doi: https://doi.org/10.1016/j.dsp.2022.103700.

[24] C. Chen, G. Wang, C. Peng, Y. Fang, D. Zhang, and H. Qin, "Exploring rich and efficient spatial-temporal interactions for real-time video salient object detection," IEEE Trans. Image Processing, vol. 30, pp. 3995-4007, Mar. 2021, doi: 10.1109/TIP.2021.3068644.

[25] H. Li, G. Chen, G. Li and Y. Yu, "Motion-guided attention for video salient object detection," ICCV, 2019, pp. 7274-7283.

[26] Z. Teed, J. Deng, "Raft: Recurrent all-pairs field transforms for optical flow," in ECCV, UK, August, 2020, pp. 402-419.

[27] R. Ranftl, A. Bochkovskiy and V. Koltun, "Vision transformers for dense prediction," ICCV, 2021, pp. 12179-12188.

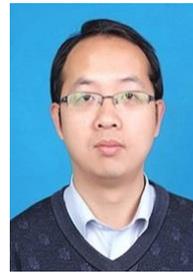

**Xiaolei Chen** received his BSc and MSc degrees from Lanzhou University, China, in 2003 and 2006 respectively, and his PhD degree from Lanzhou University of Technology in 2014, China.

He is currently an associate professor in the College of Electrical and Information Engineering at Lanzhou University of Technology. His research interests include artificial intelligence, computer vision, and virtual reality.

**Pengcheng Zhang** received his BSc degrees from Lanzhou University of Technology, China, in 2019.

He is currently a graduate student in the College of Electrical and Information Engineering at Lanzhou University of Technology, China. His research interests include virtual reality and salient object detection.

**Zelong Du** received his BSc degrees from Nanjing University of Posts and Telecommunications, China, in 2021.

He is currently a graduate student in the College of Electrical and Information Engineering at Lanzhou University of Technology, China. His research interests include virtual reality and salient object detection.

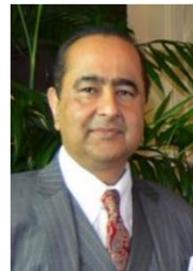

**Ishfaq Ahmad** received a B.Sc. degree in Electrical Engineering from the University of Engineering and Technology, Pakistan, in 1985, and an MS degree in Computer Engineering and a Ph.D. degree in Computer Science from Syracuse University, New York, USA, in 1987 and 1992, respectively.

Since 2002, he is a professor of Computer Science and Engineering at the University of Texas at Arlington (UTA). He has authored 270-plus publications, including books, papers in peer-reviewed journals and conference proceedings on digital video compression and analysis，parallel optimization algorithms, supercomputing systems. Google Scholar ranks him among the top ten researchers in several sub-branches of computer science and engineering.

He has served as an editor of six other journals, chaired over 20 conferences, and delivered more than 150 talks, including several keynote speeches. Dr. Ahmad is a Fellow of the IEEE.